\documentclass[10pt,twocolumn,letterpaper]{article}

\usepackage{cvpr}
\usepackage{times}
\usepackage{epsfig}
\usepackage{graphicx}
\usepackage{amsmath}
\usepackage{amssymb}

\usepackage{float}
\usepackage{booktabs}
\usepackage{colortbl}
\usepackage{xcolor}
\usepackage{subcaption}
\usepackage{multirow}
\usepackage[labelsep=period]{caption}
\RequirePackage{enumitem}
\setlist[itemize]{nosep}
\newcommand{\B}{\textbf}
\usepackage{gensymb} 

\newcommand*{\MinNumber}{0}
\newcommand*{\MidNumber}{12} 
\newcommand*{\MaxNumber}{40}
\newcommand*{\Ratio}{70}

\usepackage{pgfplots}
\usepgfplotslibrary{dateplot}
\newcommand{\cl}[1]{%
        \ifdim #1 pt > \MidNumber pt
            \pgfmathsetmacro{\PercentColor}{max(min(\Ratio*(#1 - \MidNumber)/(\MaxNumber-\MidNumber),\Ratio),0.00)} %
            \edef\x{\noexpand\cellcolor{red!\PercentColor!yellow!80}}\x #1
        \else
            \pgfmathsetmacro{\PercentColor}{max(min(\Ratio*(\MidNumber - #1)/(\MidNumber-\MinNumber),\Ratio),0.00)} %
            \edef\x{\noexpand\cellcolor{green!\PercentColor!yellow!80}}\x #1
        \fi
}

\usepackage{tikz}
\usepackage{tikz-3dplot}
\usepackage[bottom]{footmisc}
\usepackage{makecell}
\usepackage{minibox}

\newcommand{\Tref}[1]{Table~\ref{#1}}

\newcommand{\fref}[1]{Fig.~\ref{#1}}
\newcommand{\Fref}[1]{Figure~\ref{#1}}
\newcommand{\Frefs}[1]{Figures~\ref{#1}}

\newcommand{\argmin}{\mathop{\rm argmin}\limits}

\newcommand{\vn}{\boldsymbol{n}}
\newcommand{\vl}{\boldsymbol{l}}

\newcommand{\del}[1]{} 

\usepackage[pagebackref=true,breaklinks=true,letterpaper=true,colorlinks,bookmarks=false]{hyperref}

\cvprfinalcopy  

\def\cvprPaperID{1504} 

\ifcvprfinal\fi
\begin{document}

\title{Self-calibrating Deep Photometric Stereo Networks}
\author{Guanying Chen$^1$ \quad Kai Han$^2$ \quad Boxin Shi$^{3,4}$ \quad Yasuyuki Matsushita$^5$ \quad Kwan-Yee K. Wong$^1$  \vspace{0.3em}\\
$^1$The University of Hong Kong \quad$^2$University of Oxford \\
\quad$^3$Peking University \quad$^4$Peng Cheng Laboratory \quad$^5$Osaka University \\
}
\maketitle

\begin{abstract}
This paper proposes an uncalibrated photometric stereo method for non-Lambertian scenes based on deep learning. Unlike previous approaches that heavily rely on assumptions of specific reflectances and light source distributions, our method is able to determine both shape and light directions of a scene with unknown arbitrary reflectances observed under unknown varying light directions. To achieve this goal, we propose a two-stage deep learning architecture, called \emph{SDPS-Net}, which can effectively take advantage of intermediate supervision, resulting in reduced learning difficulty compared to a single-stage model. Experiments on both synthetic and real datasets show that our proposed approach significantly outperforms previous uncalibrated photometric stereo methods.
\end{abstract}

\section{Introduction}
Photometric stereo aims at recovering the surface normal of a static object from a set of images captured under different light directions~\cite{woodham1980ps,silver1980determining}. \emph{Calibrated} photometric stereo methods assume known light directions, and promising results have been reported~\cite{shi2018benchmark} at the cost of tedious light source calibration. The problem of \emph{uncalibrated} photometric stereo, where light directions are unknown, still remains an open challenge, and its stable solution is wanted because of the ease of setting. In this work, we study the problem of uncalibrated photometric stereo for surfaces with general and unknown isotropic reflectance.

Most of the existing methods for uncalibrated photometric stereo~\cite{alldrin2007r,shi2010self,papad14closed} assume a simplified reflectance model, such as the Lambertian model, and focus on resolving the shape-light ambiguity, such as the Generalized Bas-Relief (GBR) ambiguity~\cite{belhumeur1999bas}. Although methods of~\cite{lu2013uncalibrated,lu2015uncalibrated} can handle surfaces with general bidirectional reflectance distribution functions (BRDFs), they rely on a uniform distribution of light directions for deriving a solution.

Recently, with the great success of deep learning in various computer vision tasks, deep learning based methods have been introduced to calibrated photometric stereo~\cite{santo2017deep,Taniai18,ikehata2018cnn,chen2018ps}. Instead of explicitly modeling complex surface reflectances, they directly learn the mapping from reflectance observations to surface normals given light directions. Although they have obtained promising results in a calibrated setting, they cannot handle the more challenging problem of \emph{uncalibrated} photometric stereo, where light directions are unknown. One simple strategy to handle uncalibrated photometric stereo with deep learning is to directly learn the mapping from images to surface normals without taking the light directions as input. However, as reported in~\cite{chen2018ps}, the performance of such a model lags far behind those which take both images and light directions as input.

In this paper, we propose a two-stage model named Self-calibrating Deep Photometric Stereo Networks (SDPS-Net) to tackle this problem. The first stage of SDPS-Net, denoted as \emph{Lighting Calibration Network} (LCNet), takes an arbitrary number of images as input and estimates their corresponding light directions and intensities. The second stage of SDPS-Net, denoted as \emph{Normal Estimation Network} (NENet), estimates a surface normal map of a scene based on the lighting conditions estimated by LCNet and the input images. The rationales behind the design of our two-stage model are as follows. First, lighting information is very important for normal estimation since lighting is the source of various cues, such as shading and reflectance, and estimating the light directions ($3$-vectors) and intensities (scalars) is in principle much easier than directly estimating the normal map (a $3$-vector at each pixel location) together with the lighting conditions. Second, by explicitly learning to estimate light directions and intensities, the model can take advantage of the intermediate supervision, resulting in a more interpretable behavior. Last, the proposed LCNet can be seamlessly integrated with existing calibrated photometric stereo methods, which enables them to deal with unknown lighting conditions. 
Our code and model can be found at \url{https://guanyingc.github.io/SDPS-Net}.

\section{Related Work}
In this section, we review learning based photometric stereo and uncalibrated photometric stereo methods. We also briefly review the loosely related work on learning based lighting estimation. Readers are referred to~\cite{shi2018benchmark} for a comprehensive survey on calibrated photometric stereo with Lambertian surfaces and general BRDFs using non-learning based methods.

\vspace{-1.3em}
\paragraph{Learning based photometric stereo}
Recently, a few deep learning based methods have been introduced to calibrated photometric stereo~\cite{santo2017deep,Taniai18,ikehata2018cnn,chen2018ps}. Santo~\etal~\cite{santo2017deep} proposed a fully-connected network to learn the mapping from reflectance observations captured under a pre-defined set of light directions to surface normal in a pixel-wise manner. Taniai and Maehara~\cite{Taniai18} introduced an unsupervised learning framework that predicts both the surface normals and reflectance images of an object. Their model is ``trained'' at test time for each test object by minimizing the reconstruction loss between the input images and the rendered images.
Ikehata~\cite{ikehata2018cnn} introduced a fixed shape representation, called observation map, that is invariant to the number and permutation of the images. For each surface point of the object, all its observations are merged into an observation map based on the given light directions, and the observation map is then fed to a convolutional neural network (CNN) to regress the normal vector. Chen~\etal~\cite{chen2018ps} proposed a fully-convolutional network (FCN) to infer the normal map from the input image-lighting pairs, and an order-agnostic max-pooling operation was adopted to handle an arbitrary number of inputs. All the above methods assume known lighting conditions and cannot handle uncalibrated photometric stereo, where the light directions and intensities are not known a priori.

\vspace{-1.3em}
\paragraph{Uncalibrated photometric stereo}
When lighting is unknown, the surface normals of a Lambertian object can only be estimated up to a $3\times 3$ linear ambiguity~\cite{hayakawa1994photometric}, which can be reduced to a $3$-parameter GBR ambiguity~\cite{belhumeur1999bas,yuille1999determining} using the surface integrability constraint. Previous work used additional clues like albedo priors~\cite{alldrin2007r,shi2010self}, inter-reflections~\cite{chandraker2005reflections}, specular spikes~\cite{drbohlav2005can}, Torrance and Sparrow reflectance model~\cite{georghiades2003incorporating}, reflectance symmetry~\cite{tan2007isotropy,wu2013calib}, multi-view images~\cite{esteban2008multiview}, and local diffuse maxima~\cite{papad14closed}, to resolve the GBR ambiguity. Cho~\etal~\cite{cho2016photometric} considered a semi-calibrated case where the light directions are known but not their intensities. There are few works that can handle non-Lambertian surfaces under unknown lighting. Hertzmann and Seitz~\cite{hertzmann2005example} proposed an exemplar based method by inserting an additional reference object to the scene. Methods based on cues like similarity in radiance changes~\cite{sato2007shape,lu2013uncalibrated} and attached shadow~\cite{okabe2009attached} were also introduced, but they require the light sources to be uniformly distributed on the whole sphere. Recently, Lu~\etal~\cite{lu2018symps} introduced a method based on the ``constrained half-vector symmetry'' to work with non-uniform lightings. Different from these traditional methods, our method can deal with surfaces with general and unknown isotropic reflectance without the need of explicitly utilizing any additional clues or reference objects, solving a complex optimization problem at test time, or making assumptions on the light source distribution. The work most related to ours is the UPS-FCN introduced in~\cite{chen2018ps}. UPS-FCN is a single-stage model that directly regresses surface normals from images that are normalized by the known light intensities. Its performance lags far behind the calibrated methods. In contrast, our method solves the problem in two stages. We first tackles an easier problem of estimating the light directions and intensities, and then estimates the surface normals using the estimated lightings and the input images.

\vspace{-1.3em}
\paragraph{Learning based lighting estimation}
Recently, learning based single-image lighting estimation methods have attracted considerable attention. Gardner~\etal~\cite{gardner2017learning} introduced a CNN for estimating HDR environment lighting from an indoor scene image. Hold-Goeffroy~\etal~\cite{hold2017deep} learned outdoor lighting using a physically-based sky model. Weber~\etal~\cite{weber2018learning} estimated indoor environment lighting from an image of an object with known shape. Zhou~\etal~\cite{Zhou_2018_CVPR} estimated lighting, in the form of Spherical Harmonics, from a human face image by assuming a Lambertian reflectance model. Different from the above methods, our method can estimate accurate directional lightings from multiple images of a static object with general shape and non-Lambertian surface.


\begin{figure*}[t] \centering
    \includegraphics[width=\textwidth]{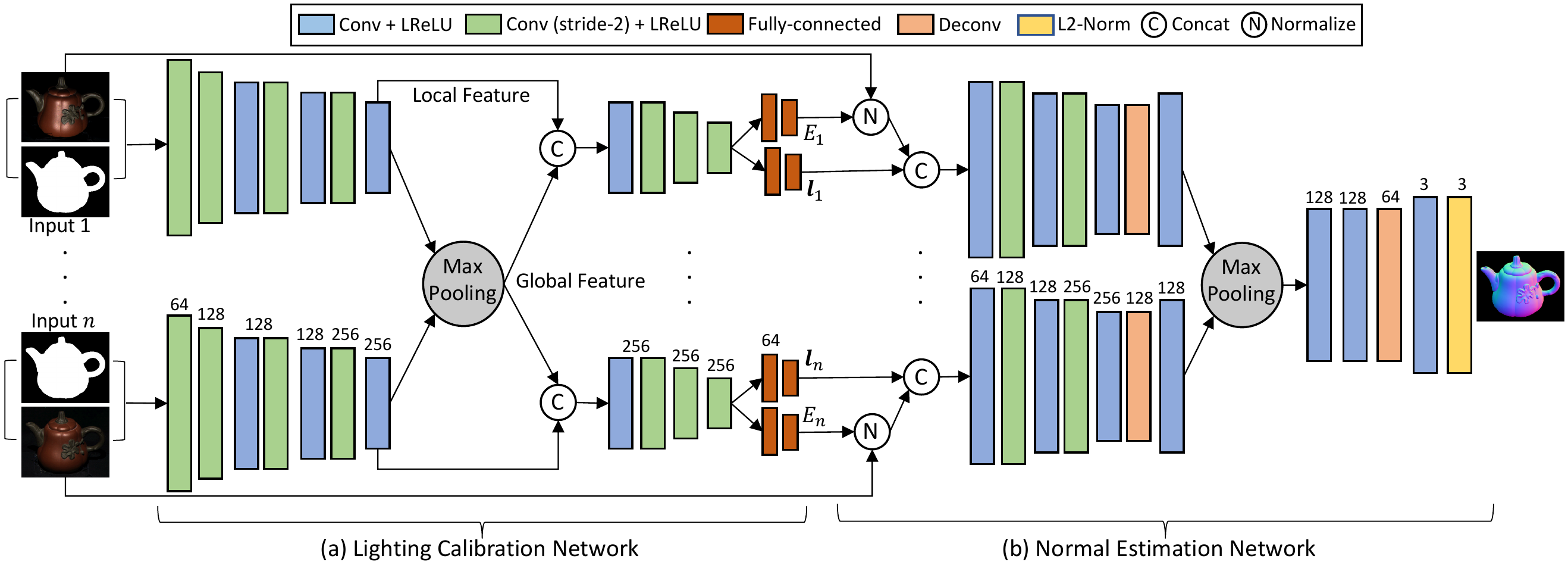}\\
    \caption{The network architecture of SDPS-Net is composed of (a) Lighting Calibration Network and (b) Normal Estimation Network. Kernel sizes for all convolutional layers are $3\times 3$, and values above the layers indicate the number of feature channels.} \label{fig:network}
\end{figure*}

\section{Image Formation Model}
Following the conventional practice, we assume an orthographic camera with linear radiometric response, white directional lightings coming from the upper-hemisphere, and the viewing direction pointing towards the viewer. 
In the rest of this paper, we refer to light direction and intensity as ``lighting''.
Consider a non-Lambertian surface whose appearance is described by a general isotropic BRDF $\rho$. Given a surface point with normal $\vn \in \mathbb{R}^3$ being illuminated by the $j$-th incoming lighting with direction $\vl_j \in \mathbb{R}^3$ and intensity $e_j \in \mathbb{R}$, the image formation model can be expressed as
\begin{equation}
    m_j = e_j \rho (\vn, \vl_j)~\text{max}(\vn^\top \vl_j, 0) + \epsilon_j,
\end{equation}
where $m$ represents the measured intensity, $\text{max}(:, 0)$ accounts for attached shadows, and $\epsilon$ accounts for the global illumination effects (cast shadows and inter-reflections) and noise.

Based on this model, given the observations of $p$ surface points under $q$ different incoming lightings, the goal of uncalibrated photometric stereo is to estimate the surface normals for these $p$ surface points given only the measured intensities.
In this work, we tackle this problem using a two-stage approach. In particular, we first estimate lightings from the measured intensities, and then solve for the surface normals using the estimated lightings and measured intensities.


\section{Learning Uncalibrated Photometric Stereo}
In this section, we introduce our two-stage framework, called SDPS-Net, for uncalibrated photometric stereo (see \fref{fig:network}). The first stage of SDPS-Net, denoted as \emph{Lighting Calibration Network} (LCNet, \fref{fig:network}~(a)), takes an arbitrary number of images as input and estimates their corresponding light directions and intensities. The second stage of SDPS-Net, denoted as \emph{Normal Estimation Network} (NENet, \fref{fig:network}~(b)), estimates an accurate normal map of the object based on the lightings estimated by LCNet and the input images.

\subsection{Lighting Calibration Network}
To estimate lightings from the images, an intuitive approach would be directly regressing the light direction vectors and intensity values.
However, we propose that formulating the lighting estimation as a classification problem is a superior choice, as will be verified by our experiments. Our arguments are as follows. Fist, classifying a light direction into a certain range is easier than regressing the exact value(s), and this will reduce the learning difficulty. Second, taking discretized light directions as input may allow NENet to better tolerate small errors in the estimated light directions.

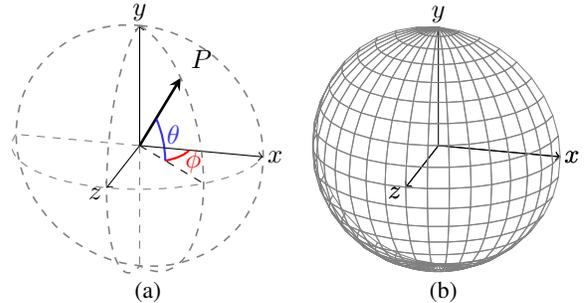
\begin{figure} \centering
    \tdplotsetmaincoords{70}{105}
\pgfmathsetmacro{\rvec}{1}
\pgfmathsetmacro{\thetavec}{40}
\pgfmathsetmacro{\phivec}{45}
\begin{tikzpicture}[scale=1.7,tdplot_main_coords]
    \coordinate (O) at (0,0,0); 
    \draw[->] (0,0,0) -- (1,0,0) node[below=0.5ex,left=-0.4ex]{$z$};
    \draw[->] (0,0,0) -- (0,1,0) node[right=-0.4ex]{$x$};
    \draw[->] (0,0,0) -- (0,0,1) node[above=-0.4ex]{$y$};
    \draw[dashed,color=gray] (0,0,0) -- (0,0,-1);
    \draw[dashed,color=gray] (0,0,0) -- (0,-1,0);
    \tdplotdrawarc[dashed,color=gray]{(O)}{1}{-90}{90}{}{}
    \tdplotsetthetaplanecoords{0}
    \tdplotdrawarc[semithick,dashed,color=gray,tdplot_rotated_coords]{(0,0,0)}{1}{0}{180}{}{}
    \tdplotsetthetaplanecoords{90}
    \tdplotdrawarc[semithick,dashed,color=gray,tdplot_rotated_coords]{(0,0,0)}{1}{0}{360}{}{}

    \tdplotsetcoord{P}{\rvec}{\thetavec}{\phivec} 
    \draw[line width=1pt,-stealth,color=black] (O) -- (P) node[above right] {$P$};

    \tdplotsetcoord{Px}{\rvec}{90}{\phivec} 
    \draw[dashed, color=black] (O) -- (Px);
    \tdplotdrawarc[thick,color=red]{(O)}{0.4}{\phivec}{90}{color=red,below=0.4ex,right=-0.4ex}{$\phi$} 

    \tdplotsetthetaplanecoords{\phivec} 
    \tdplotdrawarc[tdplot_rotated_coords,thick,color=blue,opacity=0.8]{(0,0,0)}{0.4}{\thetavec}{90}{above=0.4ex,right=-0.5ex}{$\theta$}
    \tdplotdrawarc[semithick,dashed,color=gray,tdplot_rotated_coords]{(0,0,0)}{1}{0}{180}{}{}
\end{tikzpicture}
    \raisebox{0.05\height}{\includegraphics[width=0.21\textwidth]{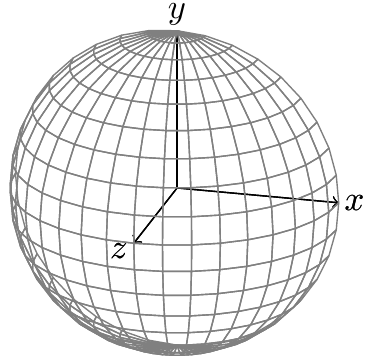}} \\ 
    \vspace{-0.9em} \makebox[0.23\textwidth]{\small (a)} \makebox[0.21\textwidth]{\small (b)} \\
    \caption{(a) Illustration of the coordinate system ($z$ axis is the viewing direction). $\phi \in [0\degree, 180\degree]$ and $\theta \in [-90\degree, 90\degree]$ are the azimuth and elevation of the light direction, respectively. (b) Example discretization of the light direction space when $K_d=18$.} \label{fig:coord}
\end{figure}

\vspace{-1.0em}
\paragraph{Discretization of lighting space} Since we cast our lighting estimation as a classification problem, we need to discretize the continuous lighting space. Note that a light direction in the upper-hemisphere can be described by its azimuth $\phi \in [0\degree, 180\degree]$ and elevation $\theta \in [-90\degree, 90\degree]$ (see \fref{fig:coord}~(a)). We can discretize the light directoin space by evenly dividing both the azimuth and elevation into $K_d$ bins, resulting in $K_d^2$ classes (see \fref{fig:coord}~(b)). Solving a $K_d^2$-class classification problem is not computationally efficient, as the softmax probability vector will have a very high dimension even when $K_d$ is not large (\eg, $K_d^2=1,296$ when $K_d=36$). 
Instead, we estimate the azimuth and elevation of a light direction separately, leading to two $K_d$-class classification problems. Similarly, we evenly divide the range of possible light intensities into $K_e$ classes (\eg, $K_e=20$ for a possible light intensity range of $[0.2, 2.0]$).

\vspace{-1.0em}
\paragraph{Local-global feature fusion}
A straightforward approach to estimate the lighting for each image is simply taking a single image as input, encoding it into a feature map using a CNN, and feeding the feature map to a lighting prediction layer. It is not surprising that the result of such a simple solution is far from satisfactory. Note that the appearance of an object is determined by its surface geometry, reflectance model and the lighting. The feature map extracted from a single observation obviously does not provide sufficient information for resolving the shape-light ambiguity. Thanks to the nature of photometric stereo where multiple observations of an object are considered, we propose a local-global feature fusion strategy to extract more comprehensive information from multiple observations.

Specifically, we separately feed each image into a shared-weight feature extractor to extract a feature map, which we call \emph{local feature} as it only provides information from a single observation. All local features of the input images are then aggregated into a \emph{global feature} through a max-pooling operation, which has been proven to be efficient and robust on aggregating salient features from a varying number of unordered inputs~\cite{wiles2017silnet,chen2018ps}. Such a global feature is expected to convey implicit surface geometry and reflectance information of the object which help resolve the ambiguity in lighting estimation. Each local feature is concatenated with the global feature, and fed to a shared-weight lighting estimation sub-network to predict the lighting for each individual image. By taking both local and global features into account, our model can produce much more reliable results than using the local features alone. We empirically found that additionally including the object mask as input can effectively improve the performance of lighting estimation, as will be seen in the experiment section.

\vspace{-1.0em}
\paragraph{Network architecture}
LCNet is a multi-input-multi-output (MIMO) network that consists of a shared-weight \emph{feature extractor}, an \emph{aggregation layer} (\ie, max-pooling layer), and a shared-weight \emph{lighting estimation sub-network} (see \fref{fig:network}~(a)). 
It takes the observations of the object together with the object mask as input, and outputs the light directions and intensities in the form of softmax probability vectors of dimension $K_d$ (azimuth), $K_d$ (elevation) and $K_e$ (intensity), respectively. 
We convert the output of LCNet to $3$-vector light directions and scalar intensity values by simply taking the middle value of the range with the highest probability\footnote{We have experimentally verified that alternative ways like taking the expectation of the probability vector or performing quadratic interpolation in the neighborhood of the peak value do not improve the result.}.

\vspace{-1.0em}
\paragraph{Loss function}
Multi-class cross entropy loss is adopted for both light direction and intensity estimation, and the overall loss function is
\begin{align}
    \label{eq:cls_loss}
    \mathcal{L}_{\text{Light}} & = \lambda_{l_a} \mathcal{L}_{l_a} + \lambda_{l_e} \mathcal{L}_{l_e} + \lambda_e \mathcal{L}_e,
\end{align}
where $\mathcal{L}_{l_a}$ and $\mathcal{L}_{l_e}$ are the loss terms for azimuth and elevation of the light direction, and  $\mathcal{L}_e$ is the loss term for light intensity. During training, weights $\lambda_{l_a}$, $\lambda_{l_e}$ and $\lambda_e$ for the loss terms are set to $1$.

\subsection{Normal Estimation Network}
NENet is a multi-input-single-output (MISO) network. The network architecture of NENet is similar to PS-FCN~\cite{chen2018ps}, consisting of a shared-weight \emph{feature extractor}, an \emph{aggregation layer}, and a \emph{normal regression sub-network} (see \fref{fig:network}~(b)). The key difference between NENet and PS-FCN is that PS-FCN requires accurate lightings as input, whereas NENet is trained with discretized lightings estimated by the LCNet and shows a more robust behavior over noise in the lightings.

NENet first normalizes the input images using the light intensities predicted by LCNet, and then concatenates the light directions predicted by LCNet with the images to form the input of the shared-weight feature extractor. Given an image of size $h\times w$, the loss function for NENet is
\begin{align}
    \label{eq:normal}
    \mathcal{L}_{\text{Normal}} = \frac{1}{hw} \sum_{i}^{hw} \left(1 - \vn_i^\top \tilde{\vn}_{i} \right),
\end{align}
where $\vn_{i}$ and $\tilde{\vn}_{i}$ denote the predicted normal and the ground-truth normal, respectively, at pixel $i$.

\subsection{Training Data}
We adopted the publicly available synthetic Blobby and Sculpture datasets~\cite{chen2018ps} for training.
Blobby and Sculpture datasets provide surfaces with complex normal distributions and diverse materials from MERL dataset~\cite{matusik2003merl}. Effects of cast shadow and inter-reflection were considered during rendering using the physically based raytracer Mitsuba~\cite{jakob2010mitsuba}. There are $85,212$ samples in total. Each sample was rendered under $64$ distinct light directions sampled from the upper-hemisphere with uniform light intensity, resulting in $5,453,568$ images ($85,212 \times 64$). The rendered images have a dimension of $128\times 128$.

To simulate images under different light intensities, we randomly generated light intensities in the range of $[0.2, 2.0]$ to scale the magnitude of the images (\ie, the ratio of the highest light intensity to the lowest one is $10$)\footnote{Note that the ratio (other than the exact value) matters, since light intensity can only be estimated up to a scale factor.}. Note that this selected range contains a wider range of intensity value than the public photometric stereo datasets like DiLiGenT benchmark~\cite{shi2018benchmark} and Gourd\&Apple dataset~\cite{alldrin2008p}. The color intensities of the input images were normalized to the range of $[0, 1]$. 
During training, we applied noise perturbation in the range of $[-0.025, 0.025]$ for data augmentation, and the input image size for LCNet and NENet was $128\times 128$. At test time, NENet can take images of different dimensions, while the input for LCNet is rescaled to $128\times 128$ as it contains fully-connected layers and requires the input to have a fixed spatial dimension. Trained only on the synthetic dataset, we will show that our model can generalize well on real datasets.

\section{Experimental Results}
We performed network analysis for our method, and compared our method with the previous state-of-the-art methods on both synthetic and real datasets.

\paragraph{Implementation details}
Our framework was implemented in PyTorch~\cite{paszke2017pytorch} and Adam optimizer~\cite{kingma2014adam} was used with default parameters.
LCNet and NENet contain $4.4$ million and $2.2$ million parameters, respectively.
We first trained LCNet using a batch size of $32$ for $20$ epochs until convergence, and then trained NENet from scratch given the lightings estimated by LCNet with a batch size of $16$ for $10$ epochs.
We found that end-to-end fine-tuning did not improve the performance.
The learning rate was initially set to $0.0005$ and halved every $5$ and $2$ epochs for LCNet and NENet, respectively.
It took about $22$ hours to train LCNet and $26$ hours to train NENet on a single Titan X Pascal GPU with a fixed input image number of $32$.

\vspace{-1.1em}
\paragraph{Evaluation metrics}
To measure the accuracy of the predicted light directions and surface normals, the widely used mean angular error (MAE) in degree is adopted.
Since the light intensities among the testing images can only be estimated up to a scale factor $s$, we introduce the scale-invariant relative error
\begin{equation}
    E_{err} = \frac{1}{q} \sum_i^q \left(\frac{|s e_i -\tilde{e}_i|}{\tilde{e}_i} \right),
\end{equation}
where $q$ is the number of images, $e_i$ and $\tilde{e}_i$ are the estimated and ground-truth light intensities, respectively, for image $i$. The scale factor $s$ is computed by solving \hbox{$\argmin_s \sum_i^n (s e_i -\tilde{e}_i)^2$} with least squares\footnote{As the calibrated intensity in the real dataset is in the form of $3$-vector, we repeat the estimated intensity to be a $3$-vector and calculate the average result.}. 

\subsection{Network Analysis with Synthetic Data}
\begin{figure} \centering
    \input{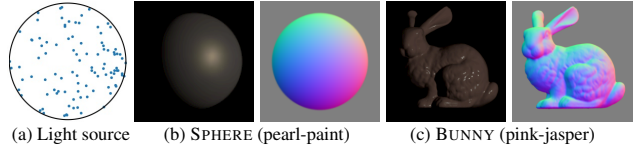}
    \caption{(a) Lighting distribution of MERL$^{\text{Test}}$ dataset. The light direction is visualized by mapping a $3$-d vector $[x,y,z]$ to a point $[x,y]$. (b) and (c) show a sample image and ground-truth normal for {\sc Sphere} and {\sc Bunny}, respectively.} \label{fig:syn_test_sample}
\end{figure}

\paragraph{MERL$^{\text{Test}}$ dataset}
To quantitatively perform network analysis for our method, we rendered a synthetic dataset, denoted as MERL$^{\text{Test}}$, of sphere and bunny shapes, denoted as {\sc Sphere} and {\sc Bunny} hereafter, respectively, using the physically based raytracer Mitsuba~\cite{jakob2010mitsuba}. Each shape was rendered with $100$ isotropic BRDFs from MERL dataset~\cite{matusik2003merl} under $100$ light directions sampled from the upper-hemisphere, leading to $200$ test objects (see \fref{fig:syn_test_sample}). Cast shadows and inter-reflections are considered for {\sc Bunny}.
For all experiments on synthetic dataset involving input with unknown light intensities, we randomly generated light intensities in the range of $[0.2, 2.0]$. Each experiment was repeated five times and the average results were reported.

\begin{figure}[t] \centering
    \tdplotsetmaincoords{70}{100}
\pgfmathsetmacro{\rvec}{1}
\pgfmathsetmacro{\thetavec}{52}
\pgfmathsetmacro{\phivec}{45}
\pgfmathsetmacro{\dev}{18}
\pgfmathsetmacro{\tha}{\thetavec-\dev}
\pgfmathsetmacro{\thb}{\thetavec+\dev}
\pgfmathsetmacro{\phia}{\phivec-\dev}
\pgfmathsetmacro{\phib}{\phivec+\dev}

\begin{tikzpicture}[scale=1.7,tdplot_main_coords]
    \coordinate (O) at (0,0,0); 
    \draw[->] (0,0,0) -- (1,0,0) node[below=0.5ex,left=-0.4ex]{$z$};
    \draw[->] (0,0,0) -- (0,1,0) node[right=-0.4ex]{$x$};
    \draw[->] (0,0,0) -- (0,0,1) node[above=-0.4ex]{$y$};
    \draw[dashed,color=gray] (0,0,0) -- (0,0,-1);
    \draw[dashed,color=gray] (0,0,0) -- (0,-1,0);
    \tdplotdrawarc[semithick,dashed,color=gray]{(O)}{1}{-90}{90}{}{}
    \tdplotsetthetaplanecoords{90}
    \tdplotdrawarc[semithick,dashed,color=gray,tdplot_rotated_coords]{(0,0,0)}{1}{0}{360}{}{}

    \tdplotsetcoord{P}{\rvec}{\thetavec}{\phivec} 
    \draw[thick,-stealth,color=black] (O) -- (P) node[below=0.2ex,right=-1.0ex] {$P$};

    \tdplotsetcoord{Pa}{\rvec}{\tha}{\phia}
    \tdplotsetcoord{Pb}{\rvec}{\tha}{\phib}
    \tdplotsetcoord{Pc}{\rvec}{\thb}{\phia}
    \tdplotsetcoord{Pd}{\rvec}{\thb}{\phib}
    \tdplotsetcoord{Pf}{\rvec}{\tha}{\phivec}

    \draw[color=black] (O) -- (Pa) node[above=0.2ex,left=-0.7ex] {\color{black!90}\scriptsize $A$};
    \draw[color=black] (O) -- (Pb) node[above,right=-0.3ex] {\color{black!90}\scriptsize $B$};
    \draw[color=black] (O) -- (Pc) node[below=1.2ex,left=-0.9ex] {\color{black!90}\scriptsize $C$};
    \draw[color=black] (O) -- (Pd) node[below=0.2ex,right=-0.3ex] {\color{black!90}\scriptsize $D$};
    \draw[color=blue,dashed] (O) -- (Pf) node[above left] {};

    \draw[semithick,color=red] (Pa) to  (Pb) node[below left] {};
    \draw[semithick,color=red] (Pd) to  (Pc) node[below left] {};

    \tdplotsetthetaplanecoords{\phivec}
    \tdplotdrawarc[thick,color=blue,opacity=0.8,tdplot_rotated_coords]{(0,0,0)}{0.7}{\tha}{\thetavec}{above=0.7ex,right=-0.7ex}{\small $\delta$}
    \tdplotsetthetaplanecoords{\phia}
    \tdplotdrawarc[semithick,color=red,tdplot_rotated_coords]{(0,0,0)}{\rvec}{\tha}{\thb}{anchor=south west}{}
    \tdplotdrawarc[dashed,tdplot_rotated_coords]{(0,0,0)}{\rvec}{0}{\tha}{anchor=south west}{}
    \tdplotdrawarc[dashed,tdplot_rotated_coords]{(0,0,0)}{\rvec}{\thb}{180}{anchor=south west}{}
    \tdplotsetthetaplanecoords{\phib}
    \tdplotdrawarc[semithick,color=red,tdplot_rotated_coords]{(0,0,0)}{\rvec}{\tha}{\thb}{anchor=south west}{}
    \tdplotdrawarc[dashed,tdplot_rotated_coords]{(0,0,0)}{\rvec}{0}{\tha}{anchor=south west}{}
    \tdplotdrawarc[dashed,tdplot_rotated_coords]{(0,0,0)}{\rvec}{\thb}{180}{anchor=south west}{}

\end{tikzpicture} \hspace{-0.5em}
    \includegraphics[width=0.25\textwidth]{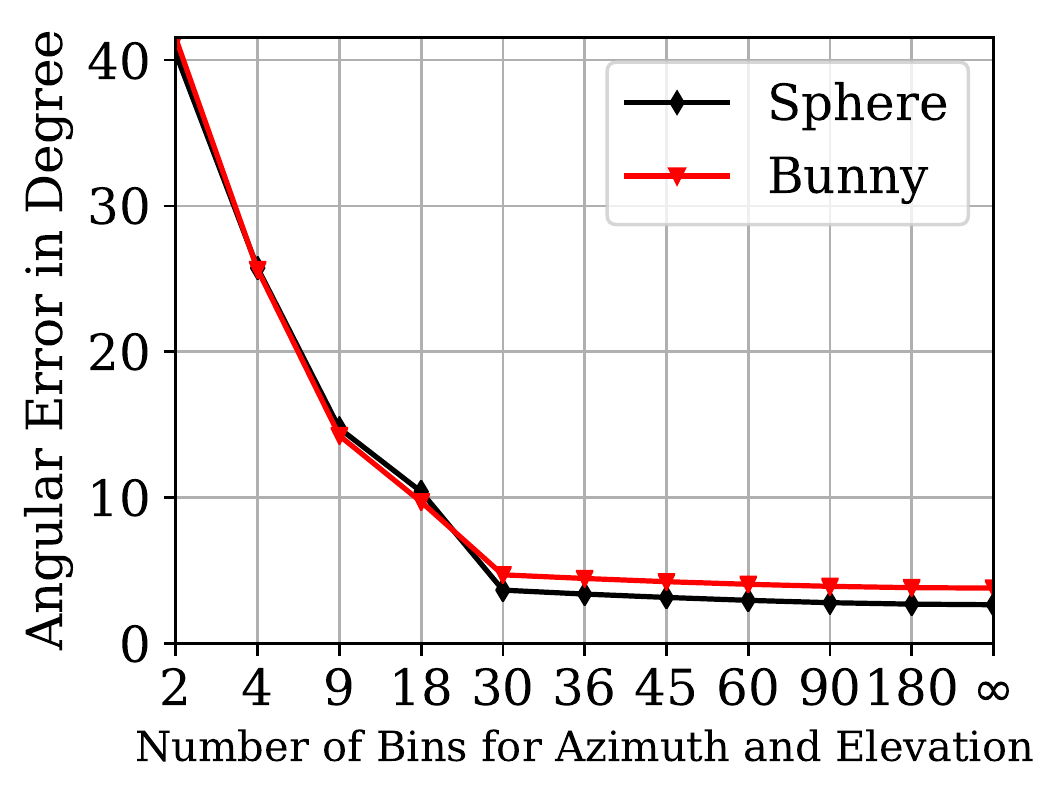}\\
    \vspace{-0.6em} \makebox[0.22\textwidth]{\small (a)} \hfill \makebox[0.23\textwidth]{\small (b)}
    \caption{(a) Light directions $A, B, C$, and $D$ have the maximum deviation angles with the light direction $P$ after discretization. (b) Upper-bound of normal estimation error for PS-FCN~\cite{chen2018ps} under different light direction space discretization levels ($\infty$ indicates no discretization).} \label{fig:discretization}
\end{figure}
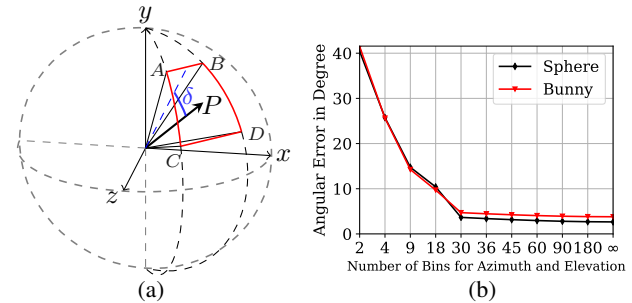

\vspace{-1.4em}
\paragraph{Discretization of lighting space}
For a given number of bins $K_d$, the maximum deviation angle for azimuth and elevation of a light direction is $\delta = 180\degree/(K_d\times 2)$ after discretization (\eg, $\delta=2.5\degree$ when $K_d=36$).
To investigate how light direction discretization affects the normal estimation accuracy, we adopted the state-of-the-art calibrated method PS-FCN~\cite{chen2018ps} and MERL$^\text{Test}$ dataset as the testbed.
We divided the azimuth and elevation of light directions into different number of bins ranging from $2$ to $180$.
For a specific bin number, we replaced each ground-truth light direction by each of the four light directions having the maximum possible angular deviations after discretization (see \fref{fig:discretization}~(a)), respectively. We then used those light directions as input for PS-FCN to infer surface normals. The normal estimation error reported in \fref{fig:discretization} (b) is the upper-bound error for PS-FCN caused by the discretization.
We can see that the increase in error caused by discretization is marginal when $K_d\ge30$.
In our implementation, we empirically set $K_d$ and $K_e$ to $36$ and $20$, respectively. We experimentally found that the performance of LCNet is robust to different discretization levels. We chose a relatively sparse discretization of lighting space in this paper as it may allow NENet to learn to better tolerate small errors in the estimated lighting at test time.

\vspace{-1.4em}
\paragraph{Effectiveness of LCNet}
To validate the design of LCNet, we compared LCNet with three baseline models for lighting estimation.
The first baseline model, denoted as LCNet$_{\text{reg}}$, is a regression based model that directly regresses the light direction vectors and intensity values (please refer to the supplementary for implementation details).
The second baseline model, denoted as LCNet$_{\text{w/o mask}}$, is a classification based model that only takes the images as input without the object mask input.
The last baseline model, denoted as LCNet$_{\text{local}}$, is a classification based model that independently estimates lighting for each observation (\ie, without local-global feature fusion).
All models were trained under the same setting, and the results are summarized in \Tref{tab:quant_light_synth}.

\begin{table}[t] \centering
    \caption{Lighting estimation results on the MERL$^\text{Test}$ dataset. The results are averaged over samples rendered with $100$ BRDFs.}
    \resizebox{0.44\textwidth}{!}{
        \begin{tabular}{c|l|*{2}{c}|*{2}{c}}
            \toprule
            & & \multicolumn{2}{c}{{\sc Sphere}} & \multicolumn{2}{c}{{\sc Bunny}} \\
            ID & Model  & Direction    & Intensity  &  Direction    & Intensity  \\
            \midrule
            \rowcolor{gray!20}
            A0 & LCNet           & \textbf{3.47} & \textbf{0.082} & \textbf{5.38}  & \textbf{0.089} \\ 
            A1 & LCNet$_{\text{reg}}$    & 4.10 & 0.104 & 5.46  & 0.094 \\ 
            A2 & LCNet$_{\text{w/o mask}}$ & 5.46 & 0.104 & 8.85 & 0.144 \\ 
            A3 & LCNet$_{\text{local}}$  & 6.87 & 0.198  & 9.98  & 0.255 \\ 
            \bottomrule
        \end{tabular}
}
 \label{tab:quant_light_synth}
\end{table}

Experiments with IDs A0 \& A1 in \Tref{tab:quant_light_synth} show that the proposed classification based LCNet consistently outperformed the regression based baseline on both light direction and intensity estimation.
This echoes our hypothesis that classifying a light direction to a certain range is easier than regressing an exact value. Thus, solving the classification problem reduces the learning difficulty and improves the performance.
Experiments with IDs A0 \& A2 show that taking the object mask as input can effectively improve the lighting estimation results.
This might be explained by the fact that object mask provides strong information for occluding contours of the object, and helps the network distinguish the shadow region from the non-object region.
Experiments with IDs A0 \& A3 show that the proposed local-global feature fusion strategy can effectively make use of information from multiple observations, and significantly improve the lighting estimation accuracy. 
Please refer to our supplementary for detailed lighting estimation results of LCNet on {\sc Bunny} from MERL$^\text{Test}$ dataset.


\vspace{-1.4em}
\paragraph{Effectiveness of NENet}
Experiments with IDs B1 \& B2 in \Tref{tab:quant_normal_syn} show that after training with the discretized lightings estimated by LCNet, NENet performs better than PS-FCN given possibly noisy lightings at test time,
while experiments with IDs B3 \& B4 show that training NENet with the light directions estimated by the regression based baseline is not always helpful.
This result further demonstrates that the proposed framework is robust to noisy lightings.
Experiments with IDs B0 \& B1 show the proposed method achieved results comparable to the fully calibrated method PS-FCN~\cite{chen2018ps}, with average MAEs of $2.71$ and $4.09$ on {\sc Sphere} and {\sc Bunny}, respectively.

\begin{table}[th] \centering
    \caption{Normal estimation results on the MERL$^\text{Test}$ dataset. The numbers are the average MAE over samples rendered with $100$ BRDFs (value the lower the better). NENet$^{\dag}$ was trained given the lightings estimated by LCNet$_{\text{reg}}$.}
    
\resizebox{0.45\textwidth}{!}{
    \begin{tabular}{c|l|c|*{1}{c}|*{1}{c}}
            \toprule
            ID & Model  & \# Param & \makebox[0.10\textwidth]{{\sc Sphere} } & \makebox[0.10\textwidth]{\sc Bunny}\\
            \midrule
            B0 & PS-FCN  \cite{chen2018ps}  & 2.2 M & 2.66   & 3.80 \\ 
            \midrule
            \rowcolor{gray!30}
            B1 & LCNet + NENet& 6.6 M & \textbf{2.71} & \textbf{4.09} \\
            B2 & LCNet + PS-FCN       & 6.6 M & 3.19  & 4.67 \\ 
            \midrule
            B3 & LCNet$_{\text{reg}}$ + NENet$^\dag$   & 6.6 M & 3.22  & 4.99 \\ 
            B4 & LCNet$_{\text{reg}}$ + PS-FCN     & 6.6 M & 3.73  & 4.96 \\
            \midrule
            B5 & UPS-FCN$_{\text{deep+mask}}$   & 6.1 M & 3.65 & 6.41 \\ 
            B6 & UPS-FCN$_{\text{deep}}$   & 6.1 M & 4.30 & 7.29 \\ 
            B7 & UPS-FCN$_{\text{wide}}$    & 6.4 M & 5.61 & 8.85 \\ 
            B8 & UPS-FCN$_{\text{est\_light}}$ & 5.7 M & 6.80 & 10.62\\
            B9 & UPS-FCN$_{\text{retrain}}$    & 2.2 M & 7.44 & 12.34 \\ 
            \bottomrule
        \end{tabular}
}
 \label{tab:quant_normal_syn}
\end{table}

\begin{figure}[t] \centering
    \includegraphics[width=0.25\textwidth]{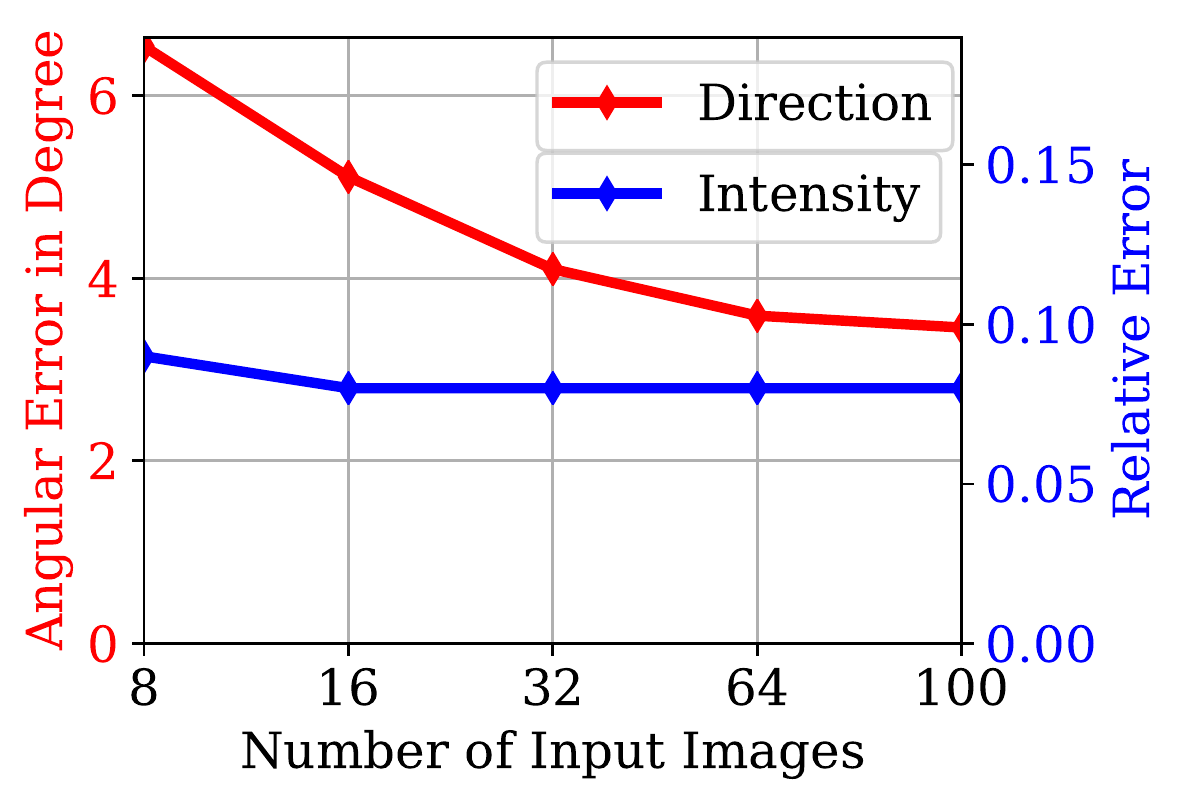}
    \includegraphics[width=0.22\textwidth]{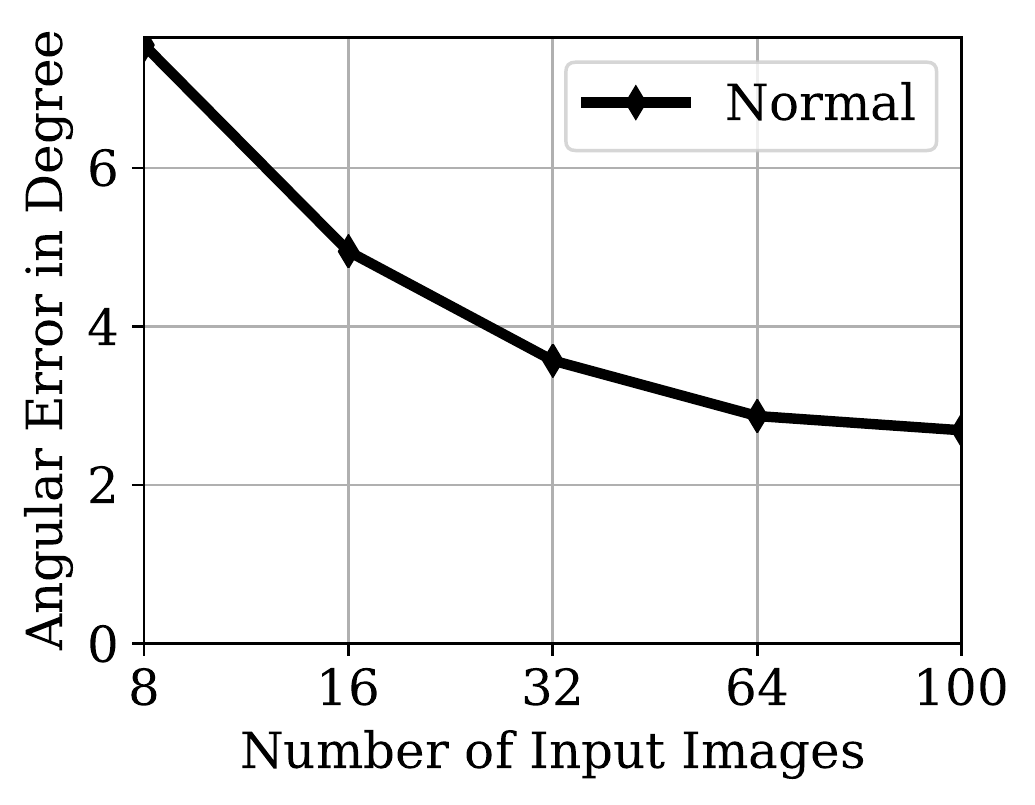}\\ 
    \caption{Results of SDPS-Net on {\sc Sphere} from MERL$^\text{Test}$ dataset with varying input image numbers.} \label{fig:img_num_syn}
\end{figure}

\Fref{fig:img_num_syn} shows that the performances of LCNet and NENet increased with the number of input images. This is expected, since more useful information can be used to infer the lightings and normals with more input images.

\vspace{-1.4em}
\paragraph{Comparison with single-stage models}
To validate the effectiveness of the proposed two-stage framework, we compared our method with five different single-stage baseline models.
We first retrained UPS-FCN~\cite{chen2018ps}, denoted as UPS-FCN$_{\text{retrain}}$, with images scaled by randomly generated light intensities to allow it adapt to unknown intensities at test time.
We then increased the model capacity of UPS-FCN by introducing a wider network (\ie, more channels in the convolutional layers) and a deeper network (\ie, more convolutional layers), denoted as UPS-FCN$_{\text{wide}}$ and UPS-FCN$_{\text{deep}}$, respectively.
We also trained a deeper network, denoted as UPS-FCN$_{\text{deep+mask}}$, that takes both the images and object mask as input. 
We last investigated the effect of having additional lighting supervision by training a variant model, denoted as UPS-FCN$_{\text{est\_light}}$, to simultaneously estimate lighting and surface normal.
Please refer to our supplementary for detailed network architectures.

\begin{figure}[t]\centering
    \input{figures/compare_quant_pf14_v2}
    \caption{Comparison between SDPS-Net and PF14~\cite{papad14closed} on {\sc Bunny} rendered with four different types of BRDFs under a near uniform lighting distribution and a biased lighting distribution.} \label{fig:compare_pf14}
\end{figure}

Experiments with IDs B5-B9 in \Tref{tab:quant_normal_syn} show that utilizing a wider or deeper network, taking the object mask as input, or incorporating additional lighting supervision can improve the performance of single-stage model in some extent.
However, experiments with IDs B1 \& B5 show that the proposed method significantly outperformed the best-performing single-stage model, especially on surfaces with complex geometry such as {\sc Bunny}, when the input as well as the number of parameters are comparable.
This result indicates that simply increasing the layer numbers or channel numbers of the network, or incorporating additional lighting supervision cannot produce optimal results. 

\vspace{-1.4em}
\paragraph{Comparison with the non-learning method~\cite{papad14closed}}
To further verify the effectiveness of our method over non-learning method, we compared SDPS-Net with the existing uncalibrated method PF14~\cite{papad14closed}, which achieved state-of-the-art results on the DiLiGenT benchmark~\cite{shi2018benchmark}, on different lighting distributions and types of BRDFs.
Specifically, we considered one near uniform and one biased lighting distribution (see \fref{fig:compare_pf14}~(a)).
We rendered {\sc Bunny} using four typical types of BRDFs, including the Lambertian model and three other types from MERL dataset~\cite{matusik2003merl}, namely, Fabric, Plastic, and Phenolic. 
They contained $15$, $12$, $9$, and $12$ different BRDFs, respectively. We reported the average results for each type (see \fref{fig:compare_pf14}~(b) for an example of each type.).

\Frefs{fig:compare_pf14}~(c)-(e) compare SDPS-Net and PF14 on lighting estimation and normal estimation. The following observations are made:
1) PF14 performed well on light direction and normal estimation for diffuse or near diffuse surfaces (\ie, Lambertian and Fabric), but will quickly degenerate when dealing with non-Lambertian surfaces. Besides, it cannot reliably estimate light intensities for all the BRDFs.
2) SDPS-Net performed well on different types of BRDFs, especially on surfaces exhibit specular highlights. This result suggests that specular highlight is an important clue for uncalibrated photometric stereo~\cite{drbohlav2005can}.
3) The performance of light direction and normal estimation of both methods will have a trend of decreasing when dealing with biased lighting distribution, while the performance of intensity estimation will slightly improve. 


\subsection{Evaluation on Real Datasets}
\begin{figure}[tbp] \centering
    \input{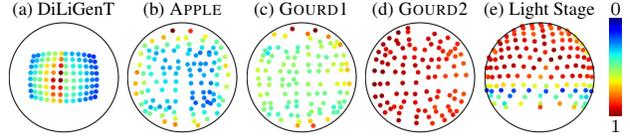}
    \caption{Lighting distributions of real testing datasets. The light direction is visualized by mapping a $3$-d vector $[x,y,z]$ to a point $[x,y]$. The color of the point indicates the light intensity (value is divided by the highest intensity to normalize to $[0,1]$).} \label{fig:light_dist}
\end{figure}

\begin{figure*}[t] \centering
    \input{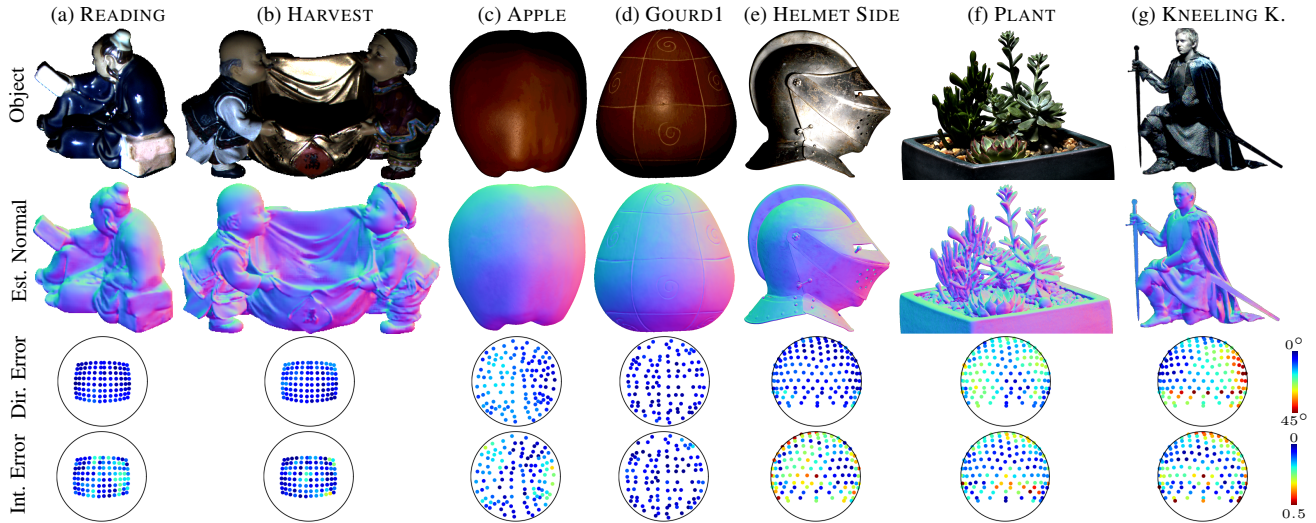}
    \caption{Qualitative results of SDPS-Net on the real testing datasets. The first to the fourth rows show the object, estimated normal map, error distribution of light direction and light intensity estimation, respectively.}
    \label{fig:qual_real_datasets}
\end{figure*}

\paragraph{Real testing datasets}
We evaluated our method on three publicly available non-Lambertian photometric stereo datasets, namely the \emph{DiLiGenT benchmark}~\cite{shi2018benchmark}, \emph{Gourd\&Apple dataset}~\cite{alldrin2008p} and \emph{Light Stage Data Gallery}~\cite{einarsson2006relighting}. \Fref{fig:light_dist} visualizes the lighting distribution of these datasets (note that for Light Stage Data Gallery, we only used $133$ images with the front side of the object under illumination).
%
%
Since Gourd\&Apple dataset and Light Stage Data Gallery only provide calibrated lightings (without ground-truth normal maps), we quantitatively evaluated our method on lighting estimation while qualitatively evaluated it on normal estimation.

\vspace{-1.2em}
\paragraph{Evaluation on DiLiGenT benchmark}
\Tref{tab:quant_light_normal_diligent} (a)-(b) show that LCNet outperformed the regression based baseline LCNet$_{\text{reg}}$ and achieved highly accurate results on both light direction and intensity estimation on DiLiGenT benchmark, with an average MAE of $4.92$ and an average relative error of $0.068$, respectively.
\Tref{tab:quant_light_normal_diligent} (c) compares the normal estimation results of SDPS-Net with previous state-of-the-art methods on DiLiGenT benchmark.
SDPS-Net achieved state-of-the-art results on almost all objects with an average MAE of $9.51$, except for the {\sc Bear} object. 
Although UPS-FCN$_\text{deep+mask}$ achieved reasonably good results on objects with smooth surface and uniform material (\eg, {\sc Ball}), it had difficulties in handling surfaces with complex geometry and spatially-varying BRDFs (\eg, {\sc Reading} and {\sc Harvest}).
The normal estimation network coupled with LCNet (i.e., SDPS-Net) outperforms that with LCNet$_{\text{reg}}$ (i.e., LCNet$_{\text{reg}}$+NENet$^\dag$) with a clear improvement of $1.52$ in average MAE, demonstrating the effectiveness of the proposed classification based LCNet.
It is interesting to see that, coupled with our LCNet, the calibrated methods L2 baseline~\cite{woodham1980ps} and IS18~\cite{ikehata2018cnn} can already achieve results comparable to the previous state-of-the-art methods. This result indicates that our proposed LCNet can be integrated with existing calibrated methods to help handle cases where lighting conditions are unknown.
\Frefs{fig:qual_real_datasets}~(a)-(b) show the qualitative results of SDPS-Net on DiLiGenT benchmark.
\begin{table}[tbp] \centering
    \caption{Results of SDPS-Net on the DiLiGenT benchmark.} \label{tab:quant_light_normal_diligent}
    \begin{subfigure}[t]{0.48\textwidth}\centering
        \caption{Results on light direction estimation.}
        \resizebox{\textwidth}{!}{
            \huge
	         \begin{tabular}{l|*{10}{c}|c}
            \toprule
 Method           & {\sc ball}         & {\sc cat}          & {\sc pot1}         & {\sc bear}         & {\sc pot2}         & {\sc buddha}        & {\sc goblet}         &{\sc reading}         & {\sc cow}            &{\sc harvest}        & Avg.\\
            \midrule
            LCNet$_{\text{reg}}$ & 4.94	 & 5.82	 & 5.62	 & 7.19	 & 4.82	 & \B{3.90}	 & 12.89	 & 7.90	 & \B{4.19}	 & 9.50	 & 6.68 \\
            LCNet & \B{3.27}	 & \B{4.08}	 & \B{5.44}	 & \B{3.47}	 & \B{2.87}	 & 4.34	 & \B{10.36}	 & \B{4.50}	 & 4.52	 & \B{6.32}	 & \B{4.92} \\
	             \bottomrule
        \end{tabular}
}

    \end{subfigure}
    \begin{subfigure}[t]{0.48\textwidth}\centering
        \caption{Results on light intensity estimation.}
        \resizebox{\textwidth}{!}{
            \huge
	         \begin{tabular}{l|*{10}{c}|c}
            \toprule
 Method           & {\sc ball}         & {\sc cat}          & {\sc pot1}         & {\sc bear}         & {\sc pot2}         & {\sc buddha}        & {\sc goblet}         &{\sc reading}         & {\sc cow}            &{\sc harvest}        & Avg.\\
            \midrule
            LCNet$_{\text{reg}}$& \B{0.032}	 & \B{0.051}	 & \B{0.048}	 & 0.167	 & 0.074	 & 0.080	 & 0.075	 & 0.141	 & \B{0.044}	 & 0.085	 & 0.080 \\
            LCNet & 0.039	 & 0.095	 & 0.058	 & \B{0.061}	 & \B{0.048}	 & \B{0.048}	 & \B{0.067}	 & \B{0.105}	 & 0.073	 & \B{0.082}	 & \B{0.068} \\
	             \bottomrule
        \end{tabular}
}

    \end{subfigure}
    \begin{subfigure}[t]{0.48\textwidth}\centering
        \caption{Results on normal estimation. (Best viewed in PDF with zoom.)} 
                \resizebox{\textwidth}{!}{
            \Huge
        \begin{tabular}{l|*{10}{c}|c}
            \toprule
            Method & {\sc ball}         & {\sc cat}          & {\sc pot1}         & {\sc bear}         & {\sc pot2}         & {\sc buddha}        & {\sc goblet}         & {\sc reading}         & {\sc cow}            &{\sc harvest}        & Avg.\\
            \midrule
            AM07~\cite{alldrin2007r}      & \cl{7.27} & \cl{31.45} & \cl{18.37} &  \cl{16.81} & \cl{49.16} & \cl{32.81} & \cl{46.54} & \cl{53.65}& \cl{54.72} & \cl{61.70} & \cl{37.25} \\
            SM10~\cite{shi2010self}       & \cl{8.90} & \cl{19.84} & \cl{16.68} &  \cl{11.98} & \cl{50.68} & \cl{15.54} & \cl{48.79} & \cl{26.93}& \cl{22.73} & \cl{73.86} & \cl{29.59} \\
            WT13~\cite{wu2013calib}       & \cl{4.39} & \cl{36.55} & \cl{9.39}  &  \textbf{\cl{6.42}}  & \cl{14.52} & \cl{13.19} & \cl{20.57} & \cl{58.96}& \cl{19.75} & \cl{55.51} & \cl{23.93} \\
            LM13~\cite{lu2013uncalibrated}& \cl{22.43}& \cl{25.01} & \cl{32.82} &  \cl{15.44} & \cl{20.57} & \cl{25.76} & \cl{29.16} & \cl{48.16}& \cl{22.53} & \cl{34.45} & \cl{27.63} \\
            PF14~\cite{papad14closed}     & \cl{4.77} & \cl{9.54}  & \cl{9.51}  &  \cl{9.07}  & \cl{15.90} & \cl{14.92} & \cl{29.93} & \cl{24.18}& \cl{19.53} & \cl{29.21} & \cl{16.66} \\
            LC18~\cite{lu2018symps}       &\cl{9.30}  &  \cl{12.60} &\cl{12.40} &\cl{10.90}	 &\cl{15.70} &\cl{19.00} &\cl{18.30} &\cl{22.30} &\cl{15.00} &\cl{28.00} &\cl{16.30} \\
UPS-FCN~\cite{chen2018ps}  &\cl{6.62}	 &\cl{14.68}	 &\cl{13.98} &\cl{11.23}	 &\cl{14.19}	 &\cl{15.87}	 &\cl{20.72}	 &\cl{23.26}	 &\cl{11.91}	 &\cl{27.79}  &\cl{16.02} \\
            \midrule
            LCNet + L2~\cite{woodham1980ps}  &\cl{4.90} &\cl{11.12} &\cl{9.72} &\cl{9.35} &\cl{14.70} &\cl{14.86} &\cl{18.29} &\cl{20.11} &\cl{25.08} &\cl{29.17} &\cl{15.73} \\
            LCNet + IS18~\cite{ikehata2018cnn}  &\cl{6.37} &\cl{15.64} &\cl{10.58} &\cl{8.48} &\cl{12.24} &\cl{13.94} &\cl{18.54} &\cl{23.78} &\cl{29.31} &\cl{25.69} &\cl{16.46} \\
            UPS-FCN$_{\text{deep+mask}}$  &\cl{3.96}	 &\cl{12.16}	 &\cl{11.13}	 &\cl{7.19}	 &\cl{11.11}	 &\cl{13.06}	 &\cl{18.07}	 &\cl{20.46}	 &\cl{11.84}	 &\cl{27.22}	 &\cl{13.62} \\
            LCNet$_{\text{reg}}$+NENet$^\dag$ &\cl{3.87}	 &\cl{8.97}	 &\textbf{\cl{8.04}}	 &\cl{15.98}	 &\cl{8.36}	 &\cl{9.42}	 &\textbf{\cl{11.49}}	 &\cl{16.99}	 &\cl{8.83}	 &\cl{18.38}	 &\cl{11.03} \\
            SDPS-Net &\textbf{\cl{2.77}}	 &\textbf{\cl{8.06}}	 &\textbf{\cl{8.14}}	 &\cl{6.89}	 &\textbf{\cl{7.50}}	 &\textbf{\cl{8.97}}	 &\textbf{\cl{11.91}}	 &\textbf{\cl{14.90}}	 &\textbf{\cl{8.48}}	 &\textbf{\cl{17.43}}	 &\textbf{\cl{9.51}} \\
            \bottomrule
        \end{tabular}
        }

    \end{subfigure}
\end{table}

\vspace{-1.2em}
\paragraph{Evaluation on other real datasets}
\Tref{tab:quant_light_gourd_stage} shows that SDPS-Net can estimate accurate light directions and intensities for the challenging Gourd\&Apple dataset and Light Stage Data Gallery.
Our method can also reliably recover visually pleasing surface normal of these two datasets (see \fref{fig:qual_real_datasets}~(c)-(g)), clearly demonstrating the practicality of the proposed methods in real world applications. 
Please refer to our supplementary for more results.

\section{Conclusion and Discussion}
In this paper, we have proposed a two-stage deep learning framework, called SDPS-Net, for uncalibrated photometric stereo. The first stage of our framework takes an arbitrary number of images as input and estimates their corresponding light directions and intensities, while the second stage predicts the normal map of the object based on the lightings estimated in the first stage and the input images.
By explicitly learning to estimate lighting conditions, our two-stage framework can take advantage of the intermediate supervision to reduce the learning difficulty and improve the final normal estimation results.
Besides, the first stage of our framework can be seamlessly integrated with existing calibrated methods, which enables them to handle uncalibrated photometric stereo.
Experiments on both synthetic and real datasets showed that our method significantly outperformed existing state-of-the-art uncalibrated photometric stereo methods.

\begin{table}[t] \centering
    \caption{Lighting estimation results of SDPS-Net on the Gourd\&Apple dataset and the Light Stage Data Gallery.}
    \label{tab:quant_light_gourd_stage}
    \begin{subfigure}[t]{0.48\textwidth}\centering
        \caption{Results on the Gourd\&Apple dataset.}
        \resizebox{0.53\textwidth}{!}{
    \Large
        \begin{tabular}{c|*{3}{c}|c}
            \toprule
            & {\sc Apple} & {\sc Gourd1} &{\sc Gourd2} & Avg.\\
            \midrule
            Direction & 9.31	 & 4.07	 & 7.11	 & 6.83  \\
            Intensity & 0.106 & 0.048& 0.186	 & 0.113 \\ 
            \bottomrule
        \end{tabular}
}

    \end{subfigure}
    \vspace{0.2em}
    \begin{subfigure}[t]{0.48\textwidth}\centering
        \caption{Results on the Light Stage Data Gallery.}
        \resizebox{\textwidth}{!}{
    \Large
        \begin{tabular}{c|*{6}{c}|c}
            \toprule
            & \makecell{{\sc Helmet} \\ {\sc Side}}& {\sc Plant} & \makecell{{\sc Fighting} \\ {\sc Knight}} & \makecell{{\sc Kneeling} \\ {\sc Knight}} & \makecell{{\sc Standing} \\ {\sc Knight}} & \makecell{{\sc Helmet}\\{\sc Front }} & Avg.\\
            \midrule
            Direction & 6.57	 & 16.06	 & 15.95	 & 19.84	 & 11.60	 & 11.62	 & 13.61 \\
            Intensity & 0.212	 & 0.170	 & 0.214	 & 0.199	 & 0.286	 & 0.248	 & 0.221 \\ 
            \bottomrule
        \end{tabular}
}

    \end{subfigure}
    \label{tab:quant_light_gourd_stage}
\end{table}

Since our framework is trained only on surfaces with uniform material, it may not perform well in dealing with steep color changes caused by multi-material surfaces (see \fref{fig:qual_real_datasets}~(b) for an example).
In the future, we will investigate better training datasets and network architectures for handling surfaces with spatially-varying BRDFs.

\vspace{-1.4em}
\paragraph{\bf Acknowledgments}
We gratefully acknowledge the support of NVIDIA Corporation with the donation of the Titan X Pascal GPU. 
Kai Han is supported by EPSRC Programme Grant Seebibyte EP/M013774/1.
Boxin Shi is supported in part by National Science Foundation of China under Grant No. 61872012. 
Yasuyuki Matsushita is supported by the New Energy and Industrial Technology Development Organization (NEDO).
{\small
\bibliographystyle{ieee_fullname}
\bibliography{gychen}
}

\clearpage
\end{document}